\documentclass[11pt,letterpaper]{article}
\usepackage[final]{nips_2016}
\usepackage{times}

\usepackage{latexsym}
\usepackage[utf8]{inputenc} 
\usepackage[T1]{fontenc}   
\usepackage{hyperref}       
\usepackage{url}            
\usepackage{booktabs}       
\usepackage{amsfonts}       
\usepackage{nicefrac}       
\usepackage{microtype}      
\usepackage[centertags]{amsmath}
\usepackage{amssymb}
\usepackage{graphicx, accents}
\usepackage{tabularx}
\usepackage{subfigure}
\usepackage{paralist}
\usepackage{authblk}
\usepackage{lipsum}

\def\x{\times}

\def\bb{\mathbf{b}}
\def\bv{\mathbf{v}}
\def\bW{\mathbf{W}}
\def\bs{\mathbf{s}}

\def\bh{\mathbf{h}}
\def\be{\mathbf{e}}
\def\bc{\mathbf{c}}
\def\bo{\mathbf{o}}

\def\mW{\mathbf{W}}

\def\Re{\mathbb{R}}

\newcommand{\sA}{\mathcal{A}}

\renewcommand{\b}[1]{\mathbf{#1}}

\newcommand{\LL}{\mathcal{L}}

\makeatletter
\DeclareRobustCommand{\cev}[1]{%
  \mathpalette\do@cev{#1}%
}
\newcommand{\do@cev}[2]{%
  \fix@cev{#1}{+}%
  \reflectbox{$\m@th#1\vec{\reflectbox{$\fix@cev{#1}{-}\m@th#1#2\fix@cev{#1}{+}$}}$}%
  \fix@cev{#1}{-}%
}
\newcommand{\fix@cev}[2]{%
  \ifx#1\displaystyle
    \mkern#23mu
  \else
    \ifx#1\textstyle
      \mkern#23mu
    \else
      \ifx#1\scriptstyle
        \mkern#22mu
      \else
        \mkern#22mu
      \fi
    \fi
  \fi
}
\makeatother

\DeclareMathOperator*{\lstm}{LSTM}

\begin{document}

\newcommand\blfootnote[1]{%
  \begingroup
  \renewcommand\thefootnote{}\footnote{#1}%
  \addtocounter{footnote}{-1}%
  \endgroup
}

\title{Machine Comprehension by Text-to-Text Neural Question Generation}

\author[1,$\dagger$]{Xingdi Yuan}
\author[1,$\dagger$]{Tong Wang}
\author[2,$\dagger$,$\ddagger$]{Caglar Gulcehre}
\author[1,$\dagger$]{Alessandro Sordoni}
\author[1]{Philip Bachman}
\author[2,$\ddagger$]{Sandeep Subramanian}
\author[2,$\ddagger$]{Saizheng Zhang}
\author[1]{Adam Trischler}
\affil[1]{Microsoft Maluuba}
\affil[ ]{{\tt\{eric.yuan,tong.wang,alsordon,phbachma,adam.trischler\}@microsoft.com}}
\affil[2]{Montreal Institute for Learning Algorithms, Universit\'e de Montr\'eal}
\affil[ ]{{\tt gulcehrc@iro.umontreal.ca} , {\tt sandeep.subramanian@gmail.com} , {\tt saizheng.zhang@umontreal.ca}}
\date{}

\maketitle
\blfootnote{$^\dagger$ These authors contributed equally.}
\blfootnote{$^\ddagger$ These authors were supported by funding from Microsoft Maluuba.}
\begin{abstract}
    We propose a recurrent neural model that generates natural-language questions from documents, conditioned on answers. We show how to train the model using a combination of supervised and reinforcement learning. After teacher forcing for standard maximum likelihood training, we fine-tune the model using policy gradient techniques to maximize several rewards that measure question quality. Most notably, one of these rewards is the performance of a question-answering system. Our model is trained and evaluated on the recent question-answering dataset \emph{SQuAD}.
\end{abstract}

\section{Introduction}
\label{sec:intro}

People ask questions to improve their knowledge and understanding of the world.
Questions can be used to access the knowledge of others or to direct one's own information-seeking behavior.
Here we study the generation of natural-language questions by machines, based on text passages.
This task is synergistic with \emph{machine comprehension} (MC), which pursues the understanding of written language by machines at a near-human level.
Because most human knowledge is recorded in text, this would enable transformative applications. 

Many machine comprehension datasets have been released recently. These generally comprise (document, question, answer) triples~\citep{hermann2015,hill2015,rajpurkar2016,trischler2016,nguyen2016}, where the goal is to predict an answer, conditioned on a document and question. The availability of large labeled datasets has spurred development of increasingly advanced models for question answering (QA) from text~\citep{kadlec2016, bidaf, mpcm, reasonet}.

\begin{table}[]
\centering
\small
\label{cherries}
\begin{tabular}{@{}p{\columnwidth}@{}}
\specialrule{.2em}{.1em}{.1em}
\textbf{Text Passage} \\
\midrule
\textbf{in \underline{1066}}$^{1,2}$ , \textbf{duke william ii}$^3$ of normandy conquered england killing king harold ii at the battle of hastings. \textbf{the invading normans and their descendants}$^4$ replaced the anglo-saxons as the ruling class of england. \\
\\
\specialrule{.2em}{.1em}{.1em}
\textbf{Questions Generated by our System} \\
\midrule
1) when did the battle of hastings take place?
\\
2) \underline{in what year was the battle of hastings fought?}
\\
3) who conquered king harold ii at the battle of hastings?
\\
4) who became the ruling class of england?\\
\specialrule{.2em}{.1em}{.1em} 
\end{tabular}
\vspace{1mm}
\caption{Examples of conditional question generation given a context and an answer from the \emph{SQuAD} dataset, using the scheme referred to as $R_{\text{PPL + QA}}$ below. Bold text in the passage indicates the answers used to generate the numbered questions.}
\end{table}

In this paper we reframe the standard MC task: rather than \emph{answering} questions about a document, we teach machines to \emph{ask} questions. Our work has several motivations.
First, we believe that posing appropriate questions is an important aspect of information acquisition in intelligent systems.
Second, learning to ask questions may improve the ability to answer them.~\citet{singer1982} demonstrated that having students devise questions before reading can increase scores on subsequent comprehension tests.
Third, answering the questions in most existing QA datasets is an \emph{extractive} task -- it requires selecting some span of text within the document -- while question asking is comparatively \emph{abstractive} -- it requires generation of text that may not appear in the document.
Fourth, asking good questions involves skills beyond those used to answer them.
For instance, in existing QA datasets, a typical (document, question) pair specifies a unique answer. Conversely, a typical (document, answer) pair may be associated with multiple questions, since a valid question can be formed from any information or relations which uniquely specify the given answer.
Finally, a mechanism to ask informative questions about documents (and eventually answer them) has many practical applications, e.g.: generating training data for question answering~\citep{serban2016generating,yang2017semi}, synthesising frequently asked question (FAQ) documentation, and automatic tutoring systems~\citep{popowich2013generating}.


We adapt the sequence-to-sequence approach of~\citet{cho2014learning} for generating questions, conditioned on a document and answer:
first we encode the document and answer, then output question words sequentially with a decoder that conditions on the document and answer encodings.
We augment the standard encoder-decoder approach with several modifications geared towards the question generation task.
During training, in addition to maximum likelihood for predicting questions from (document, answer) tuples, we use policy gradient optimization to maximize several auxiliary rewards.
These include a language-model-based score for fluency and the performance of a pretrained question-answering model on generated questions.
We show quantitatively that policy gradient increases the rewards earned by generated questions at test time, and provide examples to illustrate the qualitative effects of different training schemes.
To our knowledge, we present the first end-to-end, text-to-text model for question generation.



\section{Related Work}
\label{sec:related}

Recently, automatic question generation has received increased  attention from the research community.
It has been harnessed, for example, as a means to build automatic tutoring systems~\citep{heilman2010good,ali2010automation,popowich2013generating,labutov2015deep,mazidi2015leveraging}, to reroute queries to community question-answering systems~\citep{zhao2011}, and to enrich training data for question-answering systems~\citep{serban2016generating,yang2017semi}.

Several earlier works process documents as individual sentences using syntactic~\citep{heilman2010good,ali2010automation,kumar2015revup} or semantic-based parsing~\citep{mannem2010question,popowich2013generating}, then reformulate questions using hand-crafted rules acting on parse trees.
These traditional approaches generate questions with a high word overlap with the original text that pertain specifically to the given sentence by re-arranging the sentence parse tree.
An alternative approach is to use generic question templates whose slots can be filled with entities from the document~\citep{popowich2013generating,chali16}.
\citet{labutov2015deep}, for example, use ontology-derived templates to generate high-level questions related to larger portions of the document.
These approaches comprise pipelines of independent components that are difficult to tune for final performance measures.

More recently, neural networks have enabled end-to-end training of question generation systems.~\citet{serban2016generating} train a neural system to convert knowledge base (KB) triples into natural-language questions.
The head and the relation form a context for the question and the tail serves as the answer.
Similarly, we assume that the answer is known \emph{a priori}, but we extend the context to encompass a span of unstructured text.
\citet{mostafazadeh2016generating} use a neural architecture to generate questions from images rather than text.
Contemporaneously with this work,~\citet{yang2017semi} developed generative domain-adaptive networks, which perform question generation as an auxiliary task in training a QA system.
The main goal of their question generation is data augmentation, thus questions themselves are not evaluated.
In contrast, our work focuses primarily on developing a neural model for question generation that could be applied to a variety of downstream tasks that includes question answering.

Our model shares similarities with recent end-to-end neural QA systems,~e.g.~\citet{bidaf,mpcm}. I.e., we use an encoder-decoder structure, where the encoder processes answer and document (instead of question and document) and our decoder generates a question (instead of an answer). While existing question answering systems typically extract the answer from the document, our decoder is a fully generative model.

Finally, we relate the recent body of works that apply reinforcement learning to natural language generation, such as~\citet{lideep,ranzato2015sequence,batch2017,zhang2017}.
We similarly apply a REINFORCE-style~\citep{williams1992} algorithm to maximize various rewards earned by generated questions.


\section{Encoder-Decoder Model for Question Generation}
\label{sec:model}

We adapt the simple encoder-decoder architecture first outlined by~\citet{cho2014learning} to the question generation problem. In particular, we base our model on the attention mechanism of~\citet{bahdanau2014} and the  \emph{pointer-softmax} copying mechanism of~\citet{gulcehre2016}. In question generation, we can condition our encoder on two different sources of information (compared to the single source in neural machine translation (NMT)): a document that the question should be about and an answer that should fit the generated question. Next, we describe how we adapt the encoder and decoder architectures in detail.

\subsection{Encoder}
Our encoder is a neural model acting on two input sequences: the document, $D = (d_1, \ldots, d_n)$ and the answer, $A = (a_1, \ldots, a_m)$.
Sequence elements $d_i,~a_j \in \mathbb{R}^{D_e}$ are given by embedding vectors~\citep{bengioembed}.

In the first stage of encoding, similar to current question answering systems, e.g.~\citep{bidaf}, we augment each document word embedding with a binary feature that indicates if the document word belongs to the answer.
Then, we run a bidirectional long short-term memory~\citep{hochreiter1997} ($\lstm$) network on the augmented document sequence, producing \emph{annotation} vectors $\bh^d = (\bh^d_1, \ldots, \bh^d_n)$. Here, $\bh^d_i \in \mathbb{R}^{D_h}$ is the concatenation of the network's forward ($\vec{\bh}_i^d$) and backward hidden states ($\cev{\bh}_i^d$) for input token $i$,~i.e., $\bh^d_i = [\vec{\bh}_i^d; \cev{\bh}_i^d]$.\footnote{We use the notation $[\cdot; \cdot]$ to denote concatenation of two vectors throughout the paper.}

Our model operates on QA datasets where the answer is extractive; thus, we encode the answer $A$ using the annotation vectors corresponding to the answer word positions in the document. We assume that, without loss of generality, $A$ consists of the sequence of words $(d_s,\dots,d_e)$ in the document, s.t. $1\leq s\leq e\leq n$. We concatenate the annotation sequence $(\bh^d_s,\dots,\bh^d_e)$ with the corresponding answer word embeddings ($a_s,\dots,a_e$),~i.e., $[\bh^d_j; a_j], s\leq j \leq e$, then apply a second bidirectional $\lstm$ (bi$\lstm$) over the resulting sequence of vectors to obtain the extractive condition encoding $\bh^a \in \mathbb{R}^{D_h}$. We form $\bh^a$ by concatenating the final hidden states from each direction of the bi$\lstm$.

We also compute an initial state  $\bs_0 \in \mathbb{R}^{D_s}$ for the decoder using the annotation vectors and the extractive condition encoding:
\begin{equation*}
    \mathbf{r} = \mathbf{L}\bh^a + \frac{1}{n}\sum_i^{|D|} \bh^d_i, \hspace{2mm} \bs_0 = \tanh\left(\mathbf{W}_0\mathbf{r} + \bb_0 \right),
\end{equation*}
where $\mathbf{L} \in \mathbb{R}^{D_h \x D_h},~\bW_0 \in \mathbb{R}^{D_s \x D_h}$, and $\bb_0 \in \mathbb{R}^{D_s}$ are parameters.\footnote{Let $|X|$ denote the length of sequence $X$.}

\subsection{Decoder}
\label{sec:decoder}

Our decoder is a neural model that generates outputs $y_t$ sequentially.
At each time-step $t$, the decoder models a conditional distribution parametrized by $\theta$, 
\begin{equation}
\label{eq:p1}
    p_{\theta}(y_t | y_{<t}, D, A),
\end{equation}
where $y_{<t}$ represents the outputs at earlier time-steps.
In question generation, output $y_t$ is a word sampled according to~(\ref{eq:p1}).

When formulating questions based on documents, it is common to refer to phrases and entities that appear directly in the text.
We therefore incorporate into our decoder a mechanism for copying relevant words from $D$.
We use the pointer-softmax formulation~\citep{gulcehre2016}, which has two output layers: the \emph{shortlist} softmax and the \emph{location} softmax.
The shortlist softmax induces a distribution over words in a predefined output vocabulary.
The location softmax is a pointer network~\citep{vinyals2015} that induces a distribution  over document tokens to be copied.
A source switching network enables the model to interpolate between these distributions.


In more detail, the decoder is a recurrent neural network. Its internal state, $\bs_t \in \mathbb{R}^{D_s}$, updates according to the long short-term memory function~\citep{hochreiter1997},~i.e.,
\begin{equation}
    \bs_t = \lstm(\bs_{t-1}, y_{t-1}, \bv_t),
    \label{eq:dec}
\end{equation}
where $\bv_t$ is a the context vector computed from the document and answer encodings.

At every time-step $t$, the model computes a soft-alignment score over the document to decide which words are more relevant to the question being generated.
As in a traditional NMT architecture, the decoder computes a relevance weight $\alpha_{tj}$ for every $j$th word in the document when generating the $t$th word in the question.
Alignment score vector $\boldsymbol{\alpha}_{t} \in \Re^{|D|}$ is computed with a single layer feedforward neural network $f(\cdot)$ using the $\tanh(\cdot)$ activation function.
The scores $\boldsymbol{\alpha}_t$ are also used as the location softmax distribution.
The network defined by $f(\cdot)$ computes energies according to (\ref{eqn:energies_comp}) for the alignments, and the normalized alignments $\alpha_{tj}$ are computed as in (\ref{eqn:alpha_comp}):
\begin{align}
    e_{tj} &= f(\bh^d_j,~\bh^a,~y_t,~\bs_{t-1})  \label{eqn:energies_comp}, \\
    \alpha_{tj} &= \frac{\exp(e_{tj})}{\sum_{i=1}^T\exp(e_{ij})} \label{eqn:alpha_comp}.
\end{align}

To compute the context vector $\bv_t$ used in~(\ref{eq:dec}), we first construct context vector $\bc_t$ for the document and then concatenate it with $\bh^a$:
\begin{align}
    \bc_t &= \sum_{i=1}^{|D|} \alpha_{ti} \bh^d_i, \\
    \bv_t &= [\bc_t; \bh^a].
\end{align}

We use a deep output layer \citep{pascanu2013construct} at each time-step for the shortlist softmax vector $\bo_t$.
This layer fuses the information coming from $\bs_t$, $\bv_t$ and $y_{t-1}$ through a simple MLP to predict the word logits for the softmax as in (\ref{eqn:deep_out}).
Parameters of the softmax layer are denoted as $\mW_o \in \Re^{|V|\times D_h}$ and $\bb_o \in \Re^{|V|}$, where $|V|$ is the size of the shortlist vocabulary (2000 words herein).
\begin{align}
    \be_t &= g(\bs_t, \bv_t, y_{t-1}) \nonumber \\
    \bo_t &= \text{softmax}(\mW_o \be_t + \bb_o) \label{eqn:deep_out}
\end{align}
A source switching variable $z_t$ enables the model to interpolate between document copying and generation from shortlist. It is computed by an MLP with two hidden layers using $\tanh$ units~\citep{gulcehre2016}.
Similarly to the computation of the shortlist softmax, the switching network takes $\bs_t$, $\bv_t$ and $y_{t-1}$ as inputs.
Its output layer generates the scalar $z_t$ through the logistic sigmoid activation function.

Finally, $p_{\theta}(y_t | y_{<t}, D, A)$ is approximated by the full pointer-softmax $\b{p}_t \in \Re^{|V| + |D|}$ by concatenating $\bo_t$ and $\boldsymbol{\alpha}_t$ after both are weighted by $z_t$:
\begin{equation}
    \label{eqn:final_output}
    \b{p}_t = [z_t\bo_t ;~(1~-~z_t)\boldsymbol{\alpha}_t].
\end{equation}
As is standard in NMT, during decoding we use a beam search~\citep{graves2012} to maximize (approximately) the conditional probability of an output sequence. We discuss this in more detail in the following section.

\subsection{Training}
The model is trained initially to minimize the negative log-likelihood of the training data under the model distribution, 
\begin{equation}
    \LL = -\sum_t \log p_{\theta}(y_t |y_{<t}, D, A),
\end{equation}
where, in the decoder~as defined in (\ref{eq:dec}), the previous token $y_{t-1}$ comes from the source sequence rather than the model output (this is called \emph{teacher forcing}).

Based on our knowledge of the task, we introduce additional training signals to aid the model's learning. First we encourage the model not to generate answer words in the question.
We use the soft answer-suppression constraint given in (\ref{eqn:answer_suppression}) with the penalty hyperparameter $\lambda_s$;
$\bar{\sA}$ denotes the set of words that appear in the answer but not in the ground-truth question:
\begin{equation}
\label{eqn:answer_suppression}
\LL_s = \lambda_s \sum_t \sum_{\bar{a} \in \bar{\sA}} p_{\theta}(y_t=\bar{a} |y_{<t}, D, A).
\end{equation}
We also encourage variety in the output words to counteract the degeneracy often observed in NLG systems towards common outputs~\citep{sordoni2015}.
This is achieved with a loss term that maximizes entropy in the output softmax~(\ref{eqn:final_output}),~i.e.,
\begin{equation}
\label{eqn:entropy}
\LL_e = \lambda_e \sum_t \b{p}_t^T \log \b{p}_t.
\end{equation}

\section{Policy Gradient Optimization}
\label{sec:training}


As described above, we use teacher forcing to train our model to generate text by maximizing ground-truth likelihood.
Teacher forcing introduces critical differences between the training phase (in which the model is driven by ground-truth sequences) and the testing phase (in which the model is driven by its own outputs)~\citep{bahdanau2016}.
Significantly, teacher forcing prevents the model from making and learning from mistakes during training.
This is related to the observation that maximizing ground-truth likelihood does not teach the model how to distribute probability mass among examples other than the ground-truth, some of which may be valid questions and some of which may be completely incoherent.
This is especially problematic in language, where there are often many ways to say the same thing.
A reinforcement learning (RL) approach, by which a model is rewarded or penalized for its own actions, could mitigate these issues -- though likely at the expense of reduced stability during training.
A properly designed reward, maximized via RL, could provide a model with more information about how to distribute probability mass among sequences that do not occur in the training set \citep{norouzi2016}.

We investigate the use of RL to fine-tune our question generation model.
Specifically, we perform policy gradient optimization following a period of ``pretraining'' on maximum likelihood,
using a combination of scalar rewards correlated to question quality.
We detail this process below.
To make clear that the model is acting freely without teacher forcing, we indicate model-generated tokens with $\hat{y}_t$ and sequences with $\hat{Y}$.

\subsection{Rewards}
\paragraph{Question answering (QA)}
One obvious measure of a question's quality is whether it can be answered correctly given the context document $D$.
We therefore feed model-generated questions into a pretrained question-answering system and use that system's accuracy as a reward.
We use the recently proposed Multi-Perspective Context Matching (MPCM)~\citep{mpcm} model as our reference QA system, \emph{sans} character-level encoding.
Broadly, that model takes in a generated question $\hat{Y}$ and a document $D$, processes them through bidirectional recurrent neural networks, applies an attention mechanism, and points to the start and end tokens of the answer in $D$. After training a MPCM model on the \emph{SQuAD} dataset, the reward $R_\text{QA}(\hat{Y})$ is given by MPCM's answer accuracy on $\hat{Y}$ in terms of the F1 score, a token-based measure proposed by~\citet{rajpurkar2016} that accounts for partial word matches:
\begin{equation}
    R_\text{QA}(\hat{Y}) = \text{F1}(\hat A, A),
\end{equation}
where $\hat A = \text{MPCM}(\hat{Y})$ is the answer to the generated question by the MPCM model.
Optimizing the QA reward could lead to `friendly' questions that are either overly simplistic or that somehow cheat by exploiting quirks in the MPCM model.
One obvious way to cheat would be to inject answer words into the question.
We prevented this by masking these out in the location softmax, a hard version of the answer suppression loss (\ref{eqn:answer_suppression}).

\paragraph{Fluency (PPL)}
Another measure of quality is a question's fluency -- i.e., is it stated in proper, grammatical English? As simultaneously proposed in~\citet{zhang2017}, we use a language model to measure and reward the fluency of generated questions.
In particular, we use the perplexity assigned to $\hat{Y}$ by an $\lstm$ language model:
\begin{equation}
    R_\text{PPL}(\hat{Y}) = -2^{-\frac{1}{T}\sum_{t=1}^T \log_2 p_\text{LM}(\hat{y}_t | \hat{y}_{<t})},
\end{equation}
where the negation is to reward the model for minimizing perplexity.
The language model is trained through maximum likelihood estimation on over $80,000$ human-generated questions from \emph{SQuAD} (the training set).

\paragraph{Combination}
For the total scalar reward earned by the word sequence $\hat{Y}$, we also test a weighted combination of the individual rewards:
\begin{equation*}
    R_{\text{PPL + QA}}(\hat{Y}) = \lambda_\text{QA} R_\text{QA}(\hat{Y}) + \lambda_\text{PPL} R_\text{PPL}(\hat{Y}),
\end{equation*}
where $\lambda_\text{QA}$ and $\lambda_\text{PPL}$ are hyperparameters.
The individual reward functions use neural models to tune the neural question generator.
This is reminiscent of recent work on GANs~\citep{goodfellow2014} and actor-critic methods~\citep{bahdanau2016}.
We treat the reward models as black boxes, rather than attempting to optimize them jointly or backpropagate error signals through them.
We leave these directions for future work.

We also experimented with several other rewards, most notably the BLEU score~\citep{bleu} between $\hat{Y}$ and the ground-truth question for the given document and answer, and a softer measure of similarity between output and ground-truth based on skip-thought vectors~\citep{skip}.
Empirically, we were unable to obtain consistent improvements on these rewards through training, though this may be an issue with hyperparameter settings.

\subsection{REINFORCE}
We use the REINFORCE algorithm~\citep{williams1992} to maximize the model's expected reward.
For each generated question $\hat{Y}$, we define the loss
\begin{equation}
    \LL_\text{RL} = - \mathbb{E}_{\hat{Y}\sim\pi(\hat{Y}|D,A)}[R(\hat{Y})],
    \label{eq:rl}
\end{equation}
where $\pi$ is the policy to be trained.
The policy is a distribution over discrete actions,~i.e.~words $\hat{y}_t$ that make up the sequence $\hat{Y}$.
It is the distribution induced at the output layer of the encoder-decoder model~(\ref{eqn:final_output}), initialized with the parameters determined through likelihood optimization.\footnote{The policy also depends on the switch values but we omit these for brevity.}

REINFORCE approximates the expectation in (\ref{eq:rl}) with independent samples from the policy distribution, yielding the policy gradient
\begin{equation}
    \nabla\LL_\text{RL} \approx \sum_{t=1} \nabla \log \pi(\hat{y}_t | \hat{y}_{<t}, D, A)\frac{R(\hat{Y}) - \mu_R}{\sigma_R},
    \label{eq:pg}
\end{equation}
where the optional $\mu_R$ and $\sigma_R$ are the running mean and standard deviation of the reward, such that $R(\hat{Y})$ has zero mean and unit variance. The resulting ``whitening'' of the rewards is a simple version of PopArt~\citep{van2016learning}, and we found empirically that it stabilized learning.

It is straightforward to combine policy gradient with maximum likelihood, as both gradients can be computed by backpropagating through a properly reweighted sequence-level log-likelihood. The sequences for policy gradient are sampled from the model and weighted by a whitened reward, and the likelihood sequences are sampled from the training set and weighted by 1.

\subsection{Training Scheme}
Instead of sampling from the model's output distribution, we use beam-search to generate questions from the model and approximate the expectation in Eq.~\ref{eq:rl}. Empirically we found that rewards could not be improved through training without this approach. Randomly sampling from the model's distribution may not be as effective for estimating the modes of the generation policy and it may introduce more variance into the policy gradient.


Beam search keeps a running set of candidates that expands and contracts adaptively. At each time-step $t$, $k$ output words that maximize the probabilities of their respective paths are selected and added to the candidate sequences, where $k$ is the beam size. The probabilities of these candidates are given by their accumulated log-likelihood up to $t$.\footnote{We also experimented with a stochastic version of beam search by randomly sampling $k$ words from top-$2k$ predictions sorted by candidate sequence probability at each time step. No performance improvement was observed.}

Given a complete sample from the beam search and its accumulated log-likelihood, the gradient in (\ref{eq:pg}) can be estimated as follows. After calculating the reward with a sequence generated by beam search, we use the sample to teacher-force the decoder so as to recreate exactly the model states from which the sequence was generated. The model can then be accurately updated by coupling the parameter-independent reward with the log-likelihood of the generated sequence.
This approach adds a computational overhead but it significantly increases the initial reward values earned by the model and stabilizes policy gradient training.

We also further tune the likelihood during policy gradient optimization to prevent the model from overwriting its earlier training. We combine the policy gradient update to the model parameters, $\nabla\LL_\text{RL}$, with an update from $\nabla\LL$ based on teacher forcing on the ground-truth signal.

\section{Experiments}
\label{sec:experiments}

\subsection{Dataset}
We conducted our experiments on the \emph{SQuAD} dataset for machine comprehension~\citep{rajpurkar2016}, a large-scale, human-generated corpus of (document, question, answer) triples.
Documents are paragraphs from 536 high-PageRank Wikipedia articles covering a variety of subjects.
Questions are posed by crowdworkers in natural language and answers are spans of text in the related paragraph highlighted by the same crowdworkers. There are 107,785 question-answer pairs in total, including 87,599 training instances and 10,570 development instances.

\subsection{Baseline Seq2Seq System}
We build a simple baseline system, ``Seq2Seq,'' on the encoder-decoder architecture with attention and pointer-softmax outlined in~\citet{bahdanau2014} and~\citet{gulcehre2016}. The baseline conditions question generation on the answer by setting $\bh^a$ to be the average of the document encodings  corresponding to the answer positions in $D$.

\subsection{Quantitative Evaluation}
We use several automatic evaluation metrics to judge the quality of generated questions with respect to the ground-truth questions from the dataset. We are undertaking a large-scale human evaluation to determine how these metrics align with human judgments. The first metric is BLEU~\citep{bleu}, a standard in machine translation, which computes \{1,2,3,4\}-gram matches between generated and ground-truth questions. Next we use F1, which focuses on unigram matches~\citep{rajpurkar2016}. We also report fluency and QA performance metrics used in our reward computation. Fluency is measured by the perplexity (PPL) of the generated question computed by the pretrained question language model. The PPL score is proportional to the marginal probability $p(\hat Y)$ estimated from the corpus. The QA performance is measured by running the pretrained MPCM model on the generated questions and measuring F1 between the predicted answer and the conditioning answer.


\begin{table}[]
\centering
\small
\begin{tabular}{@{}lccccc@{}}
\toprule
                          & NLL & BLEU   & F1     & QA     & PPL      \\ \midrule \midrule
Seq2Seq                   & 45.8 & 4.9 & 31.2 & 45.6 & 153.2 \\ \midrule
Our System                & \textbf{35.3} & 10.2 & 39.5  & 65.3 & 175.7 \\
+ PG ($R_{\text{PPL}}$)      & 35.7 & 9.2 & 38.2 & 61.1 & \textbf{155.6}  \\
+ PG ($R_{\text{QA}}$)      & 39.8 & \textbf{10.5} & \textbf{40.1} & \textbf{74.2} & 300.9 \\
+ PG ($R_{\text{PPL} + \text{QA}}$) & 39.0 & 9.2 & 37.8 & 70.2 & 183.1 \\
\midrule
Question LM & - &  - & - & - &  87.7   \\
MPCM &- & - & - & 70.5 & - \\ \bottomrule
\end{tabular}
\vspace{3mm}
\caption{\label{tab:results} Automatic metrics on \emph{SQuAD}'s dev set. NLL is the negative log-likelihood. BLEU and F1 are computed with respect to the ground-truth questions. QA is the F1 obtained by the MPCM model answers to generated questions and PPL is the perplexity computed with the question language model (LM) (lower is better). PG denotes policy gradient training. The bottom two lines report performance on ground-truth questions.}
\end{table}

\begin{table}[]
\centering
\small

\begin{tabular}{@{}p{\columnwidth}@{}}
\specialrule{.2em}{.1em}{.1em}
\textbf{Text Passage} \\
\midrule
...the court of justice accepted that a requirement to speak \textbf{gaelic} to teach in a dublin design college could be justified as part of the public policy of promoting the irish language. \\
\\
\specialrule{.2em}{.1em}{.1em}
\textbf{Generated Questions}\\
\midrule
1) what did the court of justice not claim to do?
\\
\midrule
2) what language did the court of justice say should be justified as part of the public language?
\\
\midrule
3) what language did the court of justice decide to speak?
\\
\midrule
4) what language did the court of justice adopt a requirement to speak?\\
\midrule
5) what language did the court of justice say should be justified as part of?\\
\specialrule{.2em}{.1em}{.1em} 
\end{tabular}
\vspace{1mm}
\caption{\label{tab:berries} Examples of generated questions given a context and an answer. Questions are generated by the five systems in Table~\ref{tab:results}, in order.}
\end{table}

\begin{table*}[t]
\centering
\small
\label{my-label}
\begin{tabular}{@{}rp{0.8\textwidth}cc}
\specialrule{.2em}{.1em}{.1em}
Training & Generated Questions & QA & PPL  \\ \midrule

$R_{\text{PPL}}$ & what was the name of the library that was listed on the grainger market? &  0  & 73.2      \\
$R_{\text{QA}}$ & the grainger market architecture was listed in 1954 by what?  &  100  & 775      \\
$R_{\text{QA+PPL}}$ & what language did the grainger market architecture belong to?                      &  0  & 257        \\
\midrule

$R_{\text{PPL}}$ & what are the main areas of southern california?                                            &  0  & 114      \\
$R_{\text{QA}}$ & southern california is famous for what? &  16.6  & 269      \\
$R_{\text{QA+PPL}}$ & what is southern california known for?                      &  16.6  & 179        \\
\midrule

$R_{\text{PPL}}$ & what was the goal of the imperial academy of medicine? & 19.1 & 44.3 \\
$R_{\text{QA}}$ & why were confucian scholars attracted to the medical profession? & 73.7 & 405 \\
$R_{\text{QA+PPL}}$ & what did the confucian scholars believe were attracted to the medical schools? & 90.9 & 135 \\
\midrule

$R_{\text{PPL}}$ & what is an example of a theory that can be solved in theory? & 0 & 38 \\
$R_{\text{QA}}$ & in complexity theory, it is known as what? & 100 & 194 \\
$R_{\text{QA+PPL}}$ & what is an example of a theory that can cause polynomial-time solutions to be useful?                       & 100  & 37 \\

\specialrule{.2em}{.1em}{.1em}
\end{tabular}
\caption{\label{comparison} Comparison of questions from different reward combinations on the same text and answer.}
\end{table*}

\subsection{Results and qualitative analysis}
Our results for automatic evaluation on \emph{SQuAD}'s development set are presented in Table~\ref{tab:results}. Implementation details for all models are given in the supplementary material.
One striking feature is that BLEU scores are quite low for all systems tested, which relates to our earlier argument that a typical (document, answer) pair may be associated with multiple semantically-distinct questions.
This seems to be born out by the result since most generated samples look reasonable despite low BLEU scores (see Tables~\ref{cherries},~\ref{tab:berries}).

\paragraph{Our system vs. Seq2Seq} Comparing our model to the Seq2Seq baseline, we see that all metrics improve notably with the exception of PPL. Interestingly, our system performs worse in terms of PPL despite achieving lower negative log-likelihood. This, along with the improvements in BLEU, F1 and QA, suggests that our system learns a more powerful conditional model at the expense of accurately modelling the marginal distribution over questions.
It is likely challenging for the model to allocate probability mass to rarer keywords that are helpful to recover the desired answer while also minimizing perplexity.
We illustrate with samples from both models, specifically the first two samples in Table~\ref{tab:berries}. The Seq2Seq baseline generated a well-formed English question, which is also quite vague -- it is only weakly conditioned on the answer. On the other hand, our system's generated question is more specific, but still not correct given the context and perhaps less fluent given the repetition of the word \textsl{language}. We found that our proposed entropy regularization helped to avoid over-fitting and worked nicely in tandem with dropout: the training loss for our regularized model was 26.6 compared to 22.0 for the Seq2Seq baseline that used only dropout regularization.

\paragraph{Policy gradient ($R_{\text{PPL}}$:~$\lambda_\text{PPL}=0.1$)} Policy gradient training with the negative perplexity of the pretrained language model improves the generator's PPL score as desired, which approaches that of the baseline Seq2Seq model. However, QA, F1, and BLEU scores decrease. This aligns with the above observation that fluency and answerability (as measured by the automatic scores) may be in competition. 
As an example, the third sample in Table~\ref{tab:berries} is more fluent than the previous examples but does not refer to the desired answer.

\paragraph{Policy gradient ($R_{\text{QA}}$:~$\lambda_\text{QA}=1.0$)}
Policy gradient is very effective at maximizing the QA reward, gaining 8.9\% in accuracy over the improved Seq2Seq model and improving most other metrics as well. The fact that QA score is 3.7\% higher than that obtained on the ground-truth questions suggests that the question generator may have learned to exploit MPCM's answering mechanism, and the higher reported perplexity suggests questions under this scheme may be less fluent. We explore this in more detail below.
The fourth sample in Table~\ref{tab:berries}, in contrast to the others, is clearly answered by the context word \textsl{gaelic} as desired.

\paragraph{Policy gradient ($R_{\text{PPL + QA}}$:~$\lambda_\text{PPL}=0.25, \lambda_\text{QA}=0.5$)}
We attempted to improve fluency and answerability in tandem by combining QA and PPL rewards.
The PPL reward adds a prior towards questions that look natural.
According to Table~\ref{tab:results}, this optimization scheme yields a good balance of performance, improving over the maximum-likelihood model by a large margin in terms of QA performance and gaining back some PPL.
In the sample shown in Table~\ref{tab:berries}, however, the question is specific to the answer but ends prematurely.

\paragraph{} In Table~\ref{comparison} we provide additional generated samples from the different PG rewards.
This table reveals one of the `tricks' encouraged by the QA reward for improving MPCM performance: questions are often phrased with the interrogative `wh' word at the end. This gives the language high perplexity, since such questions are rarer in the training data, but brings the question form closer to the form of the source text for answer matching. 

\subsection{Discussion}
Looking through examples revealed certain difficulties in the task and some pathologies in the model that should be rectified through future work.

\paragraph{Entities and Verbs} Similar entities and related verbs are often swapped, e.g., \textsl{miami} for \textsl{jacksonville} in a question about population. This issue could be mitigated by biasing the pointer softmax towards the document for certain word types.

\paragraph{Abstraction} We desire a system that generates \emph{interesting} questions, which are not limited to reordering words from the context but exhibit some abstraction. Rewards from existing QA systems do not seem beneficial for this purpose. Questions generated through NLL training show more abstraction at the expense of decreased specificity.

\paragraph{Commonsense and Reasoning} Commonsense understanding appears critical for generating questions that are well-posed and show abstraction from the original text. Likewise, the ability to reason about and compose relations between entities could lead to more abstract and interesting questions. The existing model has no such capacity.

\begin{table}[]
\centering
\small

\begin{tabular}{@{}p{\columnwidth}@{}}
\specialrule{.2em}{.1em}{.1em}
\textbf{Text Passage} \\
\midrule
some yuan documents such as wang zhen's nong shu were printed with earthenware movable type, a technology invented in the \textbf{12th century}. \\
\\
\specialrule{.2em}{.1em}{.1em}
\textbf{Human- and Model- Generated Questions}\\
\midrule
1) when was earthenware movable type invented?
\\
\midrule
2) when was wang zhen's nong shu printed?\\
\specialrule{.2em}{.1em}{.1em} 
\end{tabular}
\vspace{1mm}
\caption{\label{tab:nongshu} A ground-truth question (1) and a valid generated question (2) with low word overlap.}
\end{table}

\paragraph{Evaluation} Due to the large number of possible questions given a predefined answer, it is challenging to evaluate the outputs using standard overlap-based metrics such as BLEU. This issue is made clear in the examples of Table~\ref{tab:nongshu}. There, the second, model-generated generated question is valid given the context and refers clearly to the answer, but has low word overlap with the first, human-generated question. This suggests that question generation from text is similar to other tasks with large output spaces~\citep{galley2015deltableu} and may benefit from corpora with multiple ground-truth questions associated to a quality rating~\citep{mostafazadeh2016generating}.



\section{Conclusion}
\label{sec:conclusion}
We proposed a recurrent neural model that generates natural-language questions conditioned on text passages and predefined answers.
We showed how to train this model using a combination of maximum likelihood and policy gradient optimization, and demonstrated both quantitatively and qualitatively how several reward combinations affect the generated outputs.
We are now undertaking a human evaluation to determine the correlation between rewards and human judgments, improving our model, and testing on additional datasets.

\bibliography{references}
\bibliographystyle{acl_natbib}

\newpage
\section*{Supplementary Material}
\appendix
\section{Implementation details}
\label{sec:impl-details}
All models are implemented using Keras \citep{keras} with Theano \citep{theano} backend. We used Adam \citep{kingmaandba2014} with an initial learning rate 2e-4 for both maximum likelihood and policy gradient updates. Word embeddings were initialized with the GloVe vectors \citep{pennington2014glove}
and updated during training. The hidden size for all RNNs is 768.

Dropout \citep{Srivastava:2014:DSW:2627435.2670313} is applied with a rate of 0.3 to the embedding layers as well as all the RNNs (between both input-hidden and hidden-hidden connections).

Both $\lambda_s$ for answer-suppression and $\lambda_e$ for entropy maximization are set to $0.01$.

We used beam search with a beam size of 32 in all experiments.

The reward weights used in policy gradient training are listed in Table~\ref{tab:reward_lambda}.

\begin{table}[h!]
\centering
\begin{tabular}{@{}rcc@{}}
\toprule
& QA$_{\text{MPCM}}$ & PPL$_{\text{Quest. LM}}$ \\
\midrule \midrule
$\lambda_{\text{QA}}$ & 1.0 & - \\
$\lambda_{\text{PPL}}$ & - & 0.1 \\
$\lambda_{\text{PPL} + \text{QA}}$ & 0.5 & 0.25 \\
\bottomrule
\end{tabular}
\vspace{2mm}
\caption{\label{tab:reward_lambda} Hyperparameter settings for policy gradient training.}
\end{table}

\end{document}